\documentclass[runningheads]{llncs}

\usepackage{amssymb}
\usepackage{graphicx}
\usepackage{url}
\usepackage{color}
\usepackage{amsfonts}
\usepackage{blindtext}
\usepackage{mathrsfs}
\usepackage{bbm}
\usepackage{url}
\usepackage{algorithm}
\usepackage{algorithmic}
\usepackage{paralist}
\usepackage[english]{babel}
\usepackage{booktabs}
\usepackage{multicol}
\usepackage{multirow}
\usepackage{subfig}
\usepackage{amsmath}

\begin{document}
\title{An Educational System for Personalized Teacher Recommendation in K-12 Online Classrooms}
%
%
 \author{Jiahao Chen \and
 Hang Li \and
 Wenbiao Ding \and
 Zitao Liu\thanks{Corresponding Author: Zitao Liu}}

 \authorrunning{J. Chen et al.}
  \titlerunning{Teacher Recommendation in K-12 Online Classrooms}
 %
 \institute{TAL Education Group, Beijing, China \\
 \email{\{chenjiahao,lihang4,dingwenbiao,liuzitao\}@tal.com}}

\maketitle              

\begin{abstract}
In this paper, we propose a simple yet effective solution to build practical teacher recommender systems for online one-on-one classes. Our system consists of (1) a pseudo matching score module that provides reliable training labels; (2) a ranking model that scores every candidate teacher; (3) a novelty boosting module that gives additional opportunities to new teachers; and (4) a diversity metric that guardrails the recommended results to reduce the chance of collision. Offline experimental results show that our approach outperforms a wide range of baselines. Furthermore, we show that our approach is able to reduce the number of student-teacher matching attempts from 7.22 to 3.09 in a five-month observation on a third-party online education platform.
\end{abstract}

\section{Introduction}
\label{sec:intro}
Because of the better accessibility and immersive learning experience, one-on-one class stands out among all the different forms of online courses \cite{chen2019multimodal,liang2005few,blatchford2011examining,blatchford2003class,finn1999tennessee}. In one-on-one courses, teacher recommender systems play an important role in helping students find their most appropriate teachers \cite{li2020identify,xu2020automatic}. However, teacher recommendation presents numerous challenges that come from the following special characteristics of real-world educational scenarios: 

\begin{itemize}

\item \textit{Limited sizes of demand and supply}: The number of teachers in supply side is incredibly smaller compared to Internet-scaled inventories. Moreover, different from item based recommendation where popular items can be suggested to millions of users simultaneously, a teacher can only take a very limited amount of students and students may only take one or two classes at each semester. 

\item \textit{Lack of gold standard}: There is no ground truth showing how good a match is between a teacher and a student. The rating based mechanism doesn't work since ratings from K-12 students are very noisy and unreliable. 

\item \textit{Cold-start teachers}: The online educational marketplace is dynamic and there are always new teachers joining. It is important to give such new teachers opportunities to let them take students instead of keeping recommending existing best performing teachers.

\item \textit{High-demand diversity}: It is undesirable to recommend the same set of teachers to students and the teacher recommender systems are supposed to reduce chances that two students want to book the same teacher at the same time.

\end{itemize}

The objective of this work is to study and develop approaches that can be used for personalized teacher recommendation for online classes. More specifically, we design techniques to (1) generate robust pseudo training labels as ground truth for learning patterns of good matches between students and teachers; (2) boost newly arrived teachers by giving incentives to their ranking scores when generating the recommended candidates; and (3) fairly evaluate and guard the diversity of recommendation results by the proposed measure of teacher diversity. We compare our approach with a wide range of baselines and evaluate its benefits on a real-world online one-on-one class dataset. We also deploy our algorithm into the real production environment and demonstrate its effectiveness in terms of number of matching attempts.






\section{The Framework}
\label{sec:method}
In this section we will discuss the details about our teacher recommendation framework for the online one-on-one courses. Our framework is made up of four key components: (1) the pseudo matching scores module; (2) the ranking model; (3) the novelty boosting module ; and (4) the diversity metric.  

\noindent \textbf{Pseudo Matching Scores} One of the most challenging problems in building teacher recommender systems is the missing of ground truth. To remedy above problem, we choose to generate the pseudo matching scores from students' dropouts. Our  mechanism relies on the assumption that matching scores reflect student preferences, which are approximated by the number of one-on-one courses between each teacher and student. In addition, we capture of the recency effect of dropout cases by using an exponential function. We design the pseudo matching scores as follows:

\begin{definition}{\textsc{positive pseudo matching score.}}
\label{def:positive}
For student $s_i$ who has completed the class, let $\mathbf{T}_i = \{t_1, t_2, \cdots, t_{p_i}\}$ be the collection of teachers who have ever taught student $s_i$ and $t_j$ denotes the $j$th teacher, $p_i$ denotes the total number of teachers who have taught student $s_i$. Let $M_i(t_j)$ be the number of courses taught by teacher $t_j$. The positive pseudo matching scores of $(s_i, t_j)$ is defined as $\mathcal{P}(s_i, t_j) = M_i(t_j) / \sum_{j=1}^{p_i} M_i(t_j)$, where $\mathcal{P}(\cdot, \cdot)$ is the positive matching score function. 
\end{definition}

\begin{definition}{\textsc{negative pseudo matching score.}}
\label{def:negative}
For student $s_k$ who has dropped the class, with similar notations in Definition \ref{def:positive}, the negative pseudo matching scores of $(s_k, t_j)$ is defined as $\mathcal{N}(s_k, t_j) = -\exp (1 - M_k(t_j)) $, where $\mathcal{N}(\cdot, \cdot)$ is the negative matching score function. 
\end{definition}

According to Definitions \ref{def:positive} and \ref{def:negative}, the pseudo matching scores range from -1 to 1. It reaches the maximum value of 1 when a student never requests a change of teacher and completes the entire class and it goes to the minimum value of -1 when a student immediately quits after the first course.

\noindent \textbf{The Ranking Model} The ranking model learns from a collection of teacher-student pairs with pseudo matching scores. We design the following three categories of features: (1) \textit{demographic features:} the demographic information of both students and teachers, such as gender, schools, etc. (2) \textit{in-class features:} the class behavioral features from both students and teachers, such as lengths of talking time, the number of spoken sentences,  etc. (3) \textit{historical features:} the historical features aggregate each teacher's past teaching performance, which includes total numbers of courses and historical dropout rates, etc. In this work, we choose to use gradient boosting decision tree (GBDT) \cite{friedman2001greedy} as our ranking model due to its robustness and generalization capability.

\noindent \textbf{Novelty Boosting for New Teachers} We design a novelty boosting component that gives extra ranking incentives to new teachers and enhances the chances of successful matches for new teachers. The novelty boosting score for teacher $t_j$ is defined as follows:

\begin{equation}
\label{eq:novelty}
b_j = \begin{cases}
\frac{\alpha}{\sqrt{ Z_j + \beta}}  & Z_j < \delta \\
0 & \text{otherwise}
\end{cases}, \quad Z_j = \sum_{i \in \mathbf{I}_j} M_i(t_j) 
\end{equation}

\noindent where $\mathbf{I}_j$ represents the index set of all taught students and $Z_j$ represents the total number of courses taught by teacher $t_j$. $\alpha$, $\beta$ and $\delta$ are positive hyper parameters. Moreover, we measure the overall effect of novelty boosting by computing the overall new teacher ratios $\label{eq:new_ratio}
r = \frac{1}{N} \sum_{i=1}^N \sum_{t_j \in \hat{\mathbf{T}}_i} (\mathbbm{1}_{b_j > 0}/|\hat{\mathbf{T}}_i|)$ in the top-recommended candidates, where $\hat{\mathbf{T}}_i$ represents the set of recommended teachers for student $s_i$ and $\mathbbm{1}_{b_j > 0}$ is the indicator function that indicates whether the teacher $t_j$ is a new teacher.

\noindent \textbf{Diversity Measurement} Diversity is important when conducting teacher recommendations in K-12 online scenarios. In this work, we propose a diversity guardrail measurement $d = 1 - \frac{2}{|\mathbf{S}|(|\mathbf{S}|-1)} \sum_{i = 1}^{|\mathbf{S}|-1} \sum_{j = i+1}^{|\mathbf{S}|} \frac{|\hat{\mathbf{T}}_i \cap \hat{\mathbf{T}}_j|}{|\mathbf{\min(|\hat{\mathbf{T}}_i|,|\hat{\mathbf{T}}_j|)}|}$, where $\mathbf{S}$ represents the set of students needed online one-on-one instructors. The diversity scores $d$ range from 0 to 1. It reaches the maximum value of 1 when each student's recommendation results don't overlap.

\section{Experiments}
\label{sec:experiment}


\noindent \textbf{Offline Evaluation}
\label{sec:offline}
The offline evaluation of recommendation is different from standard binary classification tasks where we can only partially observe the ground truth. When designing an effective offline evaluation environment, we will only focus on the ``good'' matches between students and teachers and ensure that a positively matched teacher exists in our recommended candidate list. Therefore, the performance is mainly evaluated by recall. Besides, we measure the effects of new teacher ratio and diversity. The hyper parameters in novelty boosting score function are $\alpha$=0.04, $\beta$=1, and for each student we select top 200 teachers as our recommended candidates. We collect a real-world dataset with 3,672 students, 2,139 teachers, and 8,072 student-teacher matches. Here, to simulate ``good'' matches, we first compute the pseudo matching scores for all 8,072 student-teacher matches and randomly select 821 pairs whose positive scores are over 0.5 as our test data. We compare our approach with ItemCF \cite{sarwar2001item}, SVD \cite{golub1970singular}, NMF \cite{gu2010collaborative}, DeepFM \cite{guo2017deepfm},W\&D \cite{cheng2016wide}. The results are shown in Table \ref{tab:new_teacher_ratio}. The proposed approach has competitive performance against the widely used recommendation models. Please note that the matrix factorization based baselines, such as ItemCF, SVD and NMF, cannot seamlessly integrate new teachers into their corresponding rating matrices and hence fail to recommend new teachers.

\vspace{-1cm}
\begin{table}
    \caption{Results on our offline educational dataset.}
        \label{tab:new_teacher_ratio}
    \begin{center}  
    \resizebox{0.7\textwidth}{!}{
    \begin{tabular}{lcccccc}
    \toprule
    model      & Precision       & Recall          & Diversity       & New Teacher Ratio   \\ \midrule

Our        & \textbf{0.0017} & \textbf{0.2545} & \textbf{0.7454} & \textbf{0.0333} \\
Wide\&Deep & 0.0016          & 0.2335          & 0.7446          & 0.0070          \\
DeepFM     & 0.0016          & 0.2351          & 0.7438          & 0.0013          \\
ItemCF     & 0.0014          & 0.2011          & 0.7232          & N/A          \\
SVD        & 0.0013          & 0.1909          & 0.7233          & N/A          \\
NMF        & 0.0013          & 0.1924          & 0.7233          & N/A         \\
\bottomrule
    \end{tabular}}
    \end{center}
   \vspace{-1cm}
\end{table}


\noindent \textbf{Online Experiments}
\label{sec:online}
We deploy our algorithm to a real production environment. We continuously observe the change of the mean value of the number of times a student requires to change their teacher. Over the five-month observation period (2020/01 - 2020/05), we found that the number of matching attempts decreased from 7.22 to 3.09, reflecting that our algorithm can accurately make more good recommendations to teachers.

\section{Conclusion}
\label{sec:conclusion}
In this paper, we present an end-to-end teacher recommendation framework for online one-on-one classes in the real-world scenario. The results on the real-world educational teacher recommendation dataset show that our proposed system can not only accurately recommend teachers in terms of recall but give more opportunities to new teachers in terms of new teacher ratios. Meanwhile, we guardrail the overall recommendation quality in terms of diversity experimentally. In online experiments, the proposed model is deployed in the real production system and the results show that the proposed approach is able to greatly reduce the number of matching attempts.

\section*{Acknowledgment}
This work was supported in part by National Key R\&D Program of China, under Grant No. 2020AAA0104500 and in part by Beijing Nova Program (Z201100006820068) from Beijing Municipal Science \& Technology Commission.

%
%
%
\bibliographystyle{splncs04.bst}
\bibliography{aied2021}

\begin{thebibliography}{10}
\providecommand{\url}[1]{\texttt{#1}}
\providecommand{\urlprefix}{URL }
\providecommand{\doi}[1]{https://doi.org/#1}

\bibitem{blatchford2011examining}
Blatchford, P., Bassett, P., Brown, P.: Examining the effect of class size on
  classroom engagement and teacher--pupil interaction: Differences in relation
  to pupil prior attainment and primary vs. secondary schools. Learning and
  Instruction  \textbf{21}(6),  715--730 (2011)

\bibitem{blatchford2003class}
Blatchford, P., Bassett, P., Goldstein, H., Martin, C.: Are class size
  differences related to pupils' educational progress and classroom processes?
  findings from the institute of education class size study of children aged
  5--7 years. British Educational Research Journal  \textbf{29}(5),  709--730
  (2003)

\bibitem{chen2019multimodal}
Chen, J., Li, H., Wang, W., Ding, W., Huang, G.Y., Liu, Z.: A multimodal
  alerting system for online class quality assurance. In: International
  Conference on Artificial Intelligence in Education. pp. 381--385. Springer
  (2019)

\bibitem{cheng2016wide}
Cheng, H.T., Koc, L., Harmsen, J., Shaked, T., Chandra, T., Aradhye, H.,
  Anderson, G., Corrado, G., Chai, W., Ispir, M., et~al.: Wide \& deep learning
  for recommender systems. In: Proceedings of the 1st workshop on deep learning
  for recommender systems. pp. 7--10 (2016)

\bibitem{finn1999tennessee}
Finn, J.D., Achilles, C.M.: Tennessee's class size study: Findings,
  implications, misconceptions. Educational evaluation and policy analysis
  \textbf{21}(2),  97--109 (1999)

\bibitem{friedman2001greedy}
Friedman, J.H.: Greedy function approximation: a gradient boosting machine.
  Annals of statistics pp. 1189--1232 (2001)

\bibitem{golub1970singular}
Golub, G.H., Reinsch, C.: Singular value decomposition and least squares
  solutions. Numerische mathematik  \textbf{14}(5),  403--420 (1970)

\bibitem{gu2010collaborative}
Gu, Q., Zhou, J., Ding, C.H.Q.: Collaborative filtering: Weighted nonnegative
  matrix factorization incorporating user and item graphs. In: Proceedings of
  the {SIAM} International Conference on Data Mining, {SDM} 2010, April 29 -
  May 1, 2010, Columbus, Ohio, {USA}. pp. 199--210. {SIAM} (2010).
  \doi{10.1137/1.9781611972801.18},
  \url{https://doi.org/10.1137/1.9781611972801.18}

\bibitem{guo2017deepfm}
Guo, H., Tang, R., Ye, Y., Li, Z., He, X.: Deepfm: {A} factorization-machine
  based neural network for {CTR} prediction. In: Sierra, C. (ed.) Proceedings
  of the Twenty-Sixth International Joint Conference on Artificial
  Intelligence, {IJCAI} 2017, Melbourne, Australia, August 19-25, 2017. pp.
  1725--1731. ijcai.org (2017). \doi{10.24963/ijcai.2017/239},
  \url{https://doi.org/10.24963/ijcai.2017/239}

\bibitem{li2020identify}
Li, H., Ding, W., Yang, S., Liu, Z.: Identifying at-risk {K-12} students in
  multimodal online environments: {A} machine learning approach. In:
  International Conference on Educational Data Mining (2020)

\bibitem{liang2005few}
Liang, J.K., Liu, T.C., Wang, H.Y., Chang, B., Deng, Y.C., Yang, J.C., Chou,
  C.Y., Ko, H.W., Yang, S., Chan, T.W.: A few design perspectives on one-on-one
  digital classroom environment. Journal of computer assisted learning
  \textbf{21}(3),  181--189 (2005)

\bibitem{sarwar2001item}
Sarwar, B.M., Karypis, G., Konstan, J.A., Riedl, J.: Item-based collaborative
  filtering recommendation algorithms. In: Shen, V.Y., Saito, N., Lyu, M.R.,
  Zurko, M.E. (eds.) Proceedings of the Tenth International World Wide Web
  Conference, {WWW} 10, Hong Kong, China, May 1-5, 2001. pp. 285--295. {ACM}
  (2001). \doi{10.1145/371920.372071},
  \url{https://doi.org/10.1145/371920.372071}

\bibitem{xu2020automatic}
Xu, S., Ding, W., Liu, Z.: Automatic dialogic instruction detection for k-12
  online one-on-one classes. In: International Conference on Artificial
  Intelligence in Education. pp. 340--345. Springer (2020)

\end{thebibliography}
\end{document}